# Region Convolutional Features for Multi-Label Remote Sensing Image Retrieval

Weixun Zhou[1], Xueqing Deng[2], and Zhenfeng Shao[1]

[1] State Key Laboratory of Information Engineering in Surveying, Mapping and Remote Sensing, Wuhan University, Wuhan 430079, China (email: weixunzhou1990@whu.edu.cn ; shaozhenfeng@whu.edu.cn).

[2] Electrical Engineering and Computer Science , University of California at Merced, CA 95343, USA (email: xdeng7@ucmerced.edu).

*Abstract*— **Conventional remote sensing image retrieval (RSIR) systems usually perform single-label retrieval where each image is annotated by a single label representing the most significant semantic content of the image. This assumption, however, ignores the complexity of remote sensing images, where an image might have multiple classes (i.e., multiple labels), thus resulting in worse retrieval performance. We therefore propose a novel multi-label RSIR approach with fully convolutional networks (FCN). In our approach, we first train a FCN model using a pixel-wise labeled dataset, and the trained FCN is then used to predict the segmentation maps of each image in the considered archive. We finally extract region convolutional features of each image based on its segmentation map. The region features can be either used to perform region-based retrieval or further post-processed to obtain a feature vector for similarity measure. The experimental results show that our approach achieves state-of-the-art performance in contrast to conventional single-label and recent multi-label RSIR approaches.**

## I Introduction

The recent advances in satellite technology resulted in a considerable volume of remote sensing (RS) image archives. This has presented the literature a significant challenge of searching images of interest from a large-scale RS archive. Remote sensing image retrieval (RSIR) is a simple yet effective method to solve this problem. An RSIR system generally has two main parts: 1) feature extraction in which the images are described and represented by a set of image features and 2) similarity measure in which the query image is matched with the rest images in the archive to retrieve the most similar images, but the remote sensing community has been focused mainly on developing discriminative image features due to the fact that retrieval performance greatly depends on the effectiveness of extracted features.

Conventional RSIR approaches are based on low-level visual features extracted either globally or locally. Color (spectral) features and texture features are commonly used global features for RSIR problem. In [1], the morphology-based spectral features are proposed and explored for image retrieval. In [2], morphological texture descriptors are computed and combined with bag-of-visual-words (BoVW) [3] framework to construct the feature representations for retrieval. An improved color texture descriptor is proposed by incorporating discriminative information among color channels in [4]. In contrast to global features, local features are generally extracted from image patches of interest, and often achieve better performance than global features. This is due to the fact that local representations may narrow down the semantic gap since the RS image content is characterized in a small neighborhood region [5]. As an example, the scale invariant feature transform (SIFT) descriptors are extracted and aggregated by BoVW to generate compact features for RSIR in [6]. Although these RSIR methods mentioned above can achieve reliable performance, they are essentially single-label approaches. For single-label RSIR, each image (i.e., query image and images to be retrieved) in the archive is labeled by a single, broad class label. This assumption, however, ignores the complexity of RS images, where an image might have multiple classes (i.e., multiple labels). Single labels are sufficient for RS problems with simple image classes, such as distinguishing between a building and a river, but multiple labels are required for distinguishing more complex image categories, such as dense residential and medium residential, where the differences only lie in the density of the buildings. Thus, in the case of RSIR problem with such complex image classes, multi-label RSIR approaches are needed.

To overcome the limitations of single-label RSIR methods, the remote sensing community has recently focused on developing multi-label approaches. In [7], an image scene semantic matching scheme is proposed for multi-label RSIR, in which an object-based support vector machine (SVM) classifier is used to obtain classification maps of images in the archive, and in the other work [8], image visual, object, and semantic features are combined to perform a coarse-to-fine retrieval of RS images from multiple sensors. In [9], a novel multi-label RSIR system combining spectral and spatial features are presented for hyperspectral image retrieval. In a recent work [10], semi supervised graph-theoretic method is introduced for multi-label RSIR, in which only a small number of images are manually labeled for training. These multi-label RSIR methods generally achieve better performance than single-label ones, since multiple labels can provide extra semantic information. However, the performance of these approaches (i.e., [7], [8] and [10]) depends greatly on the initial segmentation results. In addition, each image in the archive is characterized by a set of concatenated handcrafted low-level features extracted from segmented regions, while deep learning and particularly convolutional neural networks (CNN) features have been proved to be more effective on RSIR.

In this letter, we therefore propose a novel multi-label RSIR method based on fully convolutional networks (FCN). In our approach, a FCN network adapted from the pre-trained CNN is trained for segmentation and region convolutional feature extraction. In other words, the two main steps (i.e., segmentation and feature extraction) are combined in a single framework.

The rest of this letter is organized as follows. Section II presents our multi-label RSIR approach based on FCN. Section III first introduces the dataset used for training FCN and evaluating retrieval performance, and then shows the experimental results. Section IV draws some conclusions.

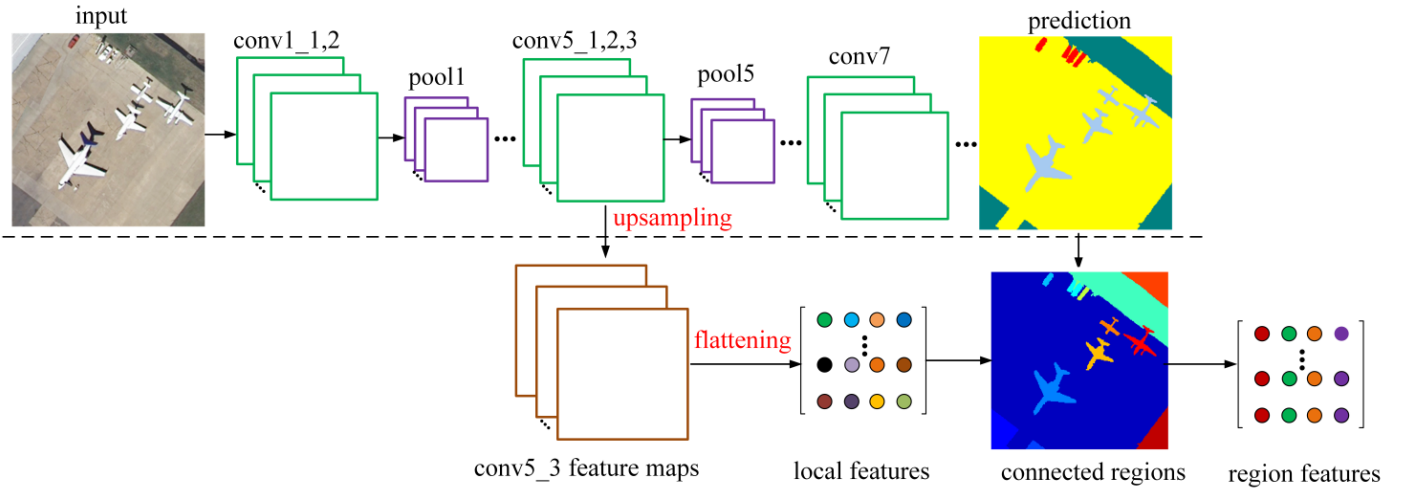

Fig.1. Flowchart of the proposed region convolutional feature for multi-label RSIR. Top: the basic architecture of FCN. conv1_1,2 and conv5_1,2,3 represent two (conv1_1, conv1_2) and three (conv5_1, conv5_2, conv5_3) convolution layers, respectively. Note that the skip connection and upsampling layers, etc. are ignored for conscience. Bottom: the process of region convolutional feature extraction. Each color in the connected regions (also called components or objects) represent one connected region.

## II Methodology

FCN has been a dominant network for dense prediction tasks like semantic segmentation since [11]. The key insight is that FCN can be trained end-to-end for pixel-wise prediction from supervised pre-training. In our work, we use the MatConvNet [12] based FCN package to build and train our FCN.

### A. Image Segmentation by FCN

In practice, FCN is usually adapted from the pre-trained CNNs such as the famous very deep network (VGG-16) [13] by transforming fully connected layers into convolution layers. For the original FCN [11], three networks (i.e., FCN-32s, FCN-16s, FCN-8s) have been trained for semantic segmentation. In this letter, we train FCN-8s (termed FCN hereafter for conscience) for segmentation and region convolutional feature extraction since it can predict finer image details, thus achieving better performance than FCN-32s and FCN-16s (We tried FCN-32s and FCN-16s in our experiments but achieved worse segmentation performance on our dataset.).

Fig.1 shows the basic architecture of our FCN. We follow the steps in [11] to build our FCN. More specially, the first two fully-connected layers of VGG-16 are converted into convolution layers (i.e., conv6 and conv7 layers in Fig.1). The last fully-connected layer (i.e., classifier layer) is modified to have 17 (the number of classes in our dataset) output classes, followed by a transposed convolution layer (also inappropriately called deconvolution layer sometimes) to upsample the coarse predictions to pixel-dense predictions. The upsampled predictions are fused with the outputs of pool3 and pool4 layers to provide further precision via skip connection layers. We refer the readers to [11] for more details on how to build FCN based on the pre-trained VGG-16 network.

### B. Region Convolutional Feature Extraction

It has been investigated that the convolutional layers of CNNs can generate local descriptors for RSIR [14]. Like SIFT, these local features also can be post-processed such as feature aggregation to generate a compact feature vector.

Fig.1 shows the process of how we extract region convolutional (termed ReCNN hereafter for conscience) features for RSIR based on the trained FCN. We first bilinearly upsample the feature maps of conv5_3 (the last convolutional layer in VGG-16) to have the same size with the images in our dataset. It is worth noting that the activation function, i.e., the rectified linear units (ReLU), is applied to these feature maps before upsampling since the use of ReLU can generate slightly better performance [13]. We then extract the local feature matrix by flattening each of these feature maps to a row feature vector, as shown in the "local features" step in Fig.1. Each column of the local feature matrix corresponds to one local descriptor and its dimension equals to the number of feature maps. The local features can be defined by:

$$f = [x_1^{(1,1)}, x_2^{(1,2)}, ..., x_m^{(p,q)}] \quad (1)$$

where $x_i^{(p,q)} (i = 1, 2, ..., m; 1 \leq p, q \leq m)$ is the local descriptor located at the pixel coordinate $(p, q)$, and $m$ is the number of local descriptors and equals to the number of image pixels. Finally, we extract the region convolutional (ReCNN) features by computing the connected regions (also called components or objects) based on the segmentation result and determining the local features that located within each of the connected regions. ReCNN can be defined by:

$$f_{ReCNN} = [y_1, y_2, ..., y_n] \quad (2)$$

where $y_j (j = 1, 2, ..., n)$ is the local descriptor of region $j$, and $n$ is the number of connected regions. To compute $y_j$, we need to determine the local descriptors $x_i^{(p,q)}$ that located within region $j$ first. This can be easily achieved by comparing the pixel coordinates between the local descriptors $x_i^{(p,q)}$ and region $j$. We then compare each dimension of these local features to obtain the maximum values (this process is also called max pooling in CNNs), which constitute the final feature vector $y_j$, as shown in "region features" step in Fig.1 (each column corresponds to a region feature vector).

After the extraction of ReCNN, each image in the archive is described by a region feature matrix. We propose two schemes to evaluate the performance of ReCNN. In the first scheme, ReCNN is used to perform a region-based retrieval. The similarity between the query image and other image in the archive are calculated by

$$D(f_q, f_r) = \frac{1}{m}\sum_{i=1}^{m}\min(D(f_i, f_j)) \quad (3)$$

where $f_q = [f_i](i = 1, 2, ..., m)$ and $f_r = [f_j](j = 1, 2, ..., n)$ are ReCNN features of the query image and other images, respectively. $D(f_i, f_j)$ is the $L_2$ distance between the region feature vector $f_i$ and $f_j$. $m$ and $n$ are the number of region feature vectors in $f_q$ and $f_r$, respectively. While in the case of the second scheme, ReCNN is further post-processed with max pooling to obtain a feature vector (termed ReCNN+ hereafter for conscience). It is worth noting that ReCNN+ here equals to applying max pooling to the local features $f$ defined in (1) directly.

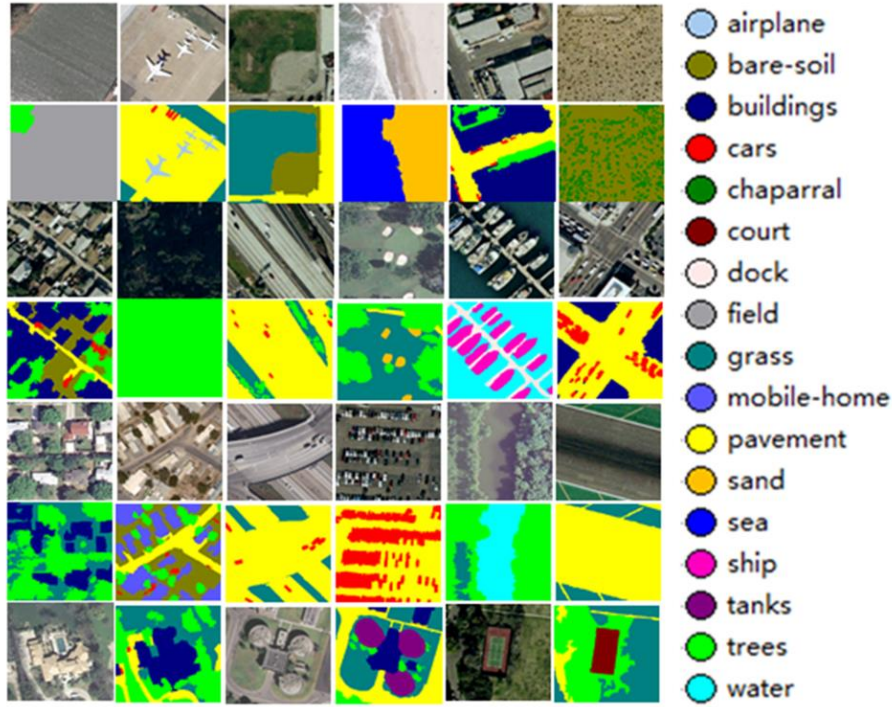

Fig.2 Example images and corresponding labeling results (The first, third, and five rows are source images, and the second, fourth, and sixth rows are corresponding labeling results, respectively.).

## III Experiments and Analysis

This section first introduces the dataset and experimental settings, and then presents the segmentation results of our FCN, the retrieval performance of ReCNN, and other state-of-the-art RSIR approaches.

### C. Dataset Description

The pixel-wise labeled dataset presented in [15] is used to train our FCN and to evaluate the performance of ReCNN and ReCNN+ features. This dataset is labeled based on the UC Merced archive [6], therefore, it also has a total number of 21 broad classes with 100 images per class. The pixels of each image in this archive are labeled with the following 17 labels, i.e., airplane, bare soil, buildings, cars, chaparral, court, dock, field, grass, mobile home, pavement, sand, sea, ship, tanks, trees, and water, which are first defined and proposed in [10]. Figure 2 shows some example images with the corresponding pixel-wise labeling results.

### D. Experimental Settings

We randomly select 80% (i.e., 1680) images as the training set to train our FCN, and the remaining 20% (i.e., 420) images are used for retrieval performance evaluation. To speed up training, the weights of VGG16 network are used as initialized weights of our FCN. We modify the last fully-connected layer of VGG16 to have 17 output classes, and initialize the new weights to zero. The weights of the transposed convolutional layer are fixed to bilinear interpolation.

In our experiments, we compare our ReCNN and ReCNN+ features with several state-of-the-arts including conventional single-label RSIR approaches such as statistics, color histogram, local binary pattern (LBP) [16], gray-level co-occurrence matrix (GLCM) [17], GIST feature [18], BoVW, and the recent multi-label RSIR approach MLIR [15]. It is worth noting that MLIR here is a bit different from its original implementation. In our implementation, we compare the region feature vectors of each image to obtain the maximum value for each dimension. Each image is then represented by a feature vector instead of a region feature matrix.

To be consistent with the work [19], we select $L_1$ distance as the distance measure for color histogram and ReCNN+, and $L_2$ for the other approaches. The average normalized modified retrieval rank (ANMRR), mean average precision (mAP), precision at $k$ (P@$k$, k is the number of returned images), and interpolated 11-points precision-recall curve, are used to evaluate the retrieval performance. For more details on these metrics, we refer the readers to [20]. We use these performance metrics for single-label RSIR to evaluate our multi-label RSIR, which makes it possible to compare our multi-label approaches with those single-label ones. It is worth noting that for ANMRR, the lower values indicate better performance, while for mAP and P@$k$, the larger the better. In addition, the final retrieval results are the averaged performance of 420 queries, and the query image itself is also viewed as a similar image in our experiments.

*E. Experimental Results*

ReCNN and ReCNN+ features of each image are extracted based on its semantic segmentation result, as described in Section II. Therefore, it is necessary to evaluate the segmentation performance of our FCN network. The mean IU, pixel accuracy, and mean accuracy values achieved by our FCN are 0.6761, 0.8056, and 0.8232, respectively. We refer the readers to [11] for more details on these three segmentation metrics. In our experiments, we also tried FCN-16s and FCN-32s networks but achieved worse performance. Fig.3 shows some segmentation results (without any post-processing) achieved by our FCN and corresponding ground truth images. It can be observed that our FCN successfully divide different land use/land cover classes into different segments.

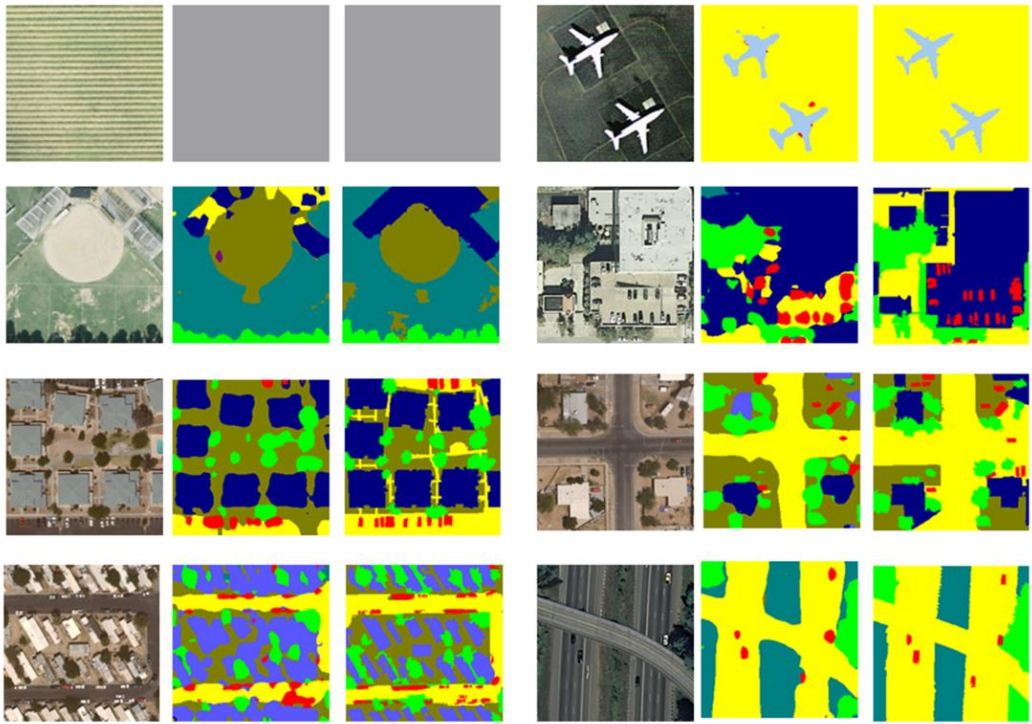

Fig.3 The segmentation results achieved by FCN (For each row, the first and last three images are source images, segmentation results, and ground truth, respectively.).

Table I
Comparisons between multi-label approach and state-of-the-art RSIR methods.

| Features | ANMRR | mAP | P@5 | P@10 | P@20 | P@50 |
|---|---|---|---|---|---|---|
| Statistics | 0.820 | 0.156 | 0.273 | 0.182 | 0.131 | 0.098 |
| Color | 0.705 | 0.255 | 0.481 | 0.341 | 0.239 | 0.146 |
| LBP | 0.740 | 0.217 | 0.480 | 0.327 | 0.218 | 0.121 |
| GLCM | 0.746 | 0.207 | 0.400 | 0.279 | 0.196 | 0.129 |
| GIST | 0.754 | 0.225 | 0.451 | 0.303 | 0.200 | 0.120 |
| BoVW | 0.538 | 0.398 | 0.561 | 0.464 | 0.376 | 0.236 |
| MLIR | 0.707 | 0.246 | 0.394 | 0.289 | 0.229 | 0.156 |
| ReCNN | 0.509 | 0.441 | 0.686 | 0.556 | 0.414 | 0.228 |
| ReCNN+ | **0.264** | **0.688** | **0.861** | **0.753** | **0.624** | **0.344** |

ReCNN and ReCNN+ features are compared with several state-of-the-art RSIR approaches, and the results are shown in Table I. We can see ReCNN and particularly ReCNN+ outperform those single-label RSIR approaches, i.e., statistics feature, color histogram, LBP, GLCM, GIST, and BoVW by a significant margin in terms of ANMRR, mAP and P@$k$ values. For example, the ANMRR values achieved by BoVW (the best performing single-label method), ReCNN, and ReCNN+ are 0.538, 0.509 and 0.264, respectively. With respect to the multi-label RSIR approach, i.e., MLIR, ReCNN and ReCNN+ also result in better performance. This is due to the fact that CNN features are more powerful than those handcrafted features, and thus are suitable for RSIR problem. It is worth noting that multi-label RSIR approaches usually achieve better performance than those single-label ones, however, MLIR performs worse than BoVW, as shown in Table I. A possible explanation is that MLIR is based on the concatenation of low-level global features, while local features have proved to be more effective for RSIR problem. An interesting finding is ReCNN+ improves ReCNN by ~25% in terms of ANMRR value. This is due to the fact that ReCNN performs region-based retrieval that ignores the spatial relationship between different regions.

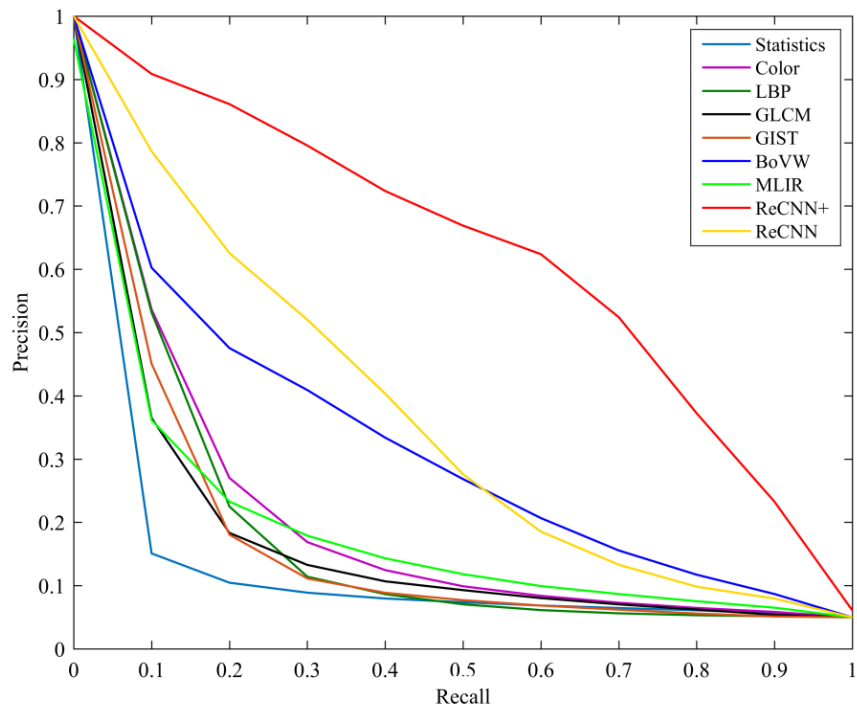

Fig.4. The precision-recall curves for our ReCNN and other state-of-the-art RSIR methods.

We plot the interpolated 11-points precision-recall graphs to further evaluate the performance of these RSIR approaches, as shown in Fig.4. The results here are consistent with that shown in Table I, i.e., ReCNN and particularly ReCNN+ improves the retrieval performance of conventional single-label and recent multil-label (i.e., MLIR) RSIR approaches substantially.

## IV Conclusion

In this letter, we present a novel multi-label RSIR approach based on FCN. In our approach, a FCN network is trained to predict the segmentation results of the images in our retrieval archive. We then upsample the feature maps and determine the local features that located within each of the connected regions to extract region convolutional features. Finally, two schemes are proposed to evaluate the performance of the proposed region convolutional features.

Experimental results show that our approach improves the retrieval performance of the state-of-the-art RSIR approaches including conventional single-label and recent multi-label methods. It is worth noting that the retrieval performance of our proposed approach (the first scheme, i.e., ReCNN) can be further improved if the spatial relationships between different regions are considered when computing image similarity. In addition, the multiple labels of each image are only considered during FCN training and region feature extraction, however, previous works have shown that multiple labels can be used to perform a coarse retrieval to filter out images that do not have overlapped classes with the query images, thus improving the search efficiency and retrieval performance.

In our future work, we are planning to extract multi-scale region convolutional features, and take the spatial relationships between different connected regions into consideration when computing image similarity.